\documentclass[pdflatex,sn-mathphys-num]{sn-jnl}
\usepackage{graphicx}
\usepackage[utf8]{inputenc}
\usepackage{multirow}
\usepackage{amsmath,amssymb,amsfonts}
\usepackage{amsthm}
\usepackage{mathrsfs}
\usepackage[title]{appendix}
\usepackage{xcolor}
\usepackage{textcomp}
\usepackage{manyfoot}
\usepackage{booktabs}
\usepackage{algorithm}
\usepackage{algorithmicx}
\usepackage{algpseudocode}
\usepackage{listings}
\usepackage{tabularx} 

\usepackage[acronym]{glossaries}
\newacronym{co}{CO}{Course Outcome}
\newacronym{po}{PO}{Program Outcome}
\newacronym{pso}{PSO}{Program-Specific Outcome}
\newacronym{rnn}{RNN}{Recurrent Neural Network}
\newacronym{cam}{CAM}{Course Articulation Matrix}
\newacronym{lime}{LIME}{Local Interpretable Model-agnostic Explanations}
\newacronym{lstm}{LSTM}{Long Short-Term Memory network}
\newacronym{svm}{SVM}{Support Vector Machine}
\newacronym{nlp}{NLP}{Natural Language Processing}
\newacronym{cbit}{CBIT}{Chaitanya Bharathi Institute of Technology}

\UseRawInputEncoding

\begin{document}

\title{BERT-Based Approach for Automating Course Articulation Matrix Construction with Explainable AI}

\author[1]{\fnm{Natenaile Asmamaw} \sur{Shiferaw}}\email{2101020423@cgu-odisha.ac.in}
\equalcont{These authors contributed equally to this work.}

\author[1]{\fnm{Simpenzwe Honore} \sur{Leandre}}\email{2101020234@cgu-odisha.ac.in}
\equalcont{These authors contributed equally to this work.}

\author[1]{\fnm{Aman} \sur{Sinha}}\email{2101020397@cgu-odisha.ac.in}

\author*[2]{\fnm{Dillip} \sur{Rout}}\email{dillip.rout@cgu-odisha.ac.in}

\affil[1]{\orgdiv{Student, Dept. of Computer Science and Engineering}, \orgname{C.V. Raman Global University}, \orgaddress{\street{Bhubaneswar}, \postcode{752054}, \state{Odisha}, \country{India}}}
\affil*[2]{\orgdiv{Faculty, Dept. of Computer Science and Engineering}, \orgname{C.V. Raman Global University}, \orgaddress{\street{Bhubaneswar}, \postcode{752054}, \state{Odisha}, \country{India}}}

\abstract{\gls{co} and \gls{po}/\gls{pso} alignment is a crucial task for ensuring curriculum coherence and assessing educational effectiveness. The construction of a \gls{cam}, which quantifies the relationship between \gls{co}s and \gls{po}s/\gls{pso}s, typically involves assigning numerical values (0, 1, 2, 3) to represent the degree of alignment. In this study, We experiment with four models from the BERT family: BERT Base, DistilBERT, ALBERT, and RoBERTa, and use multiclass classification to assess the alignment between \gls{co} and \gls{po}/\gls{pso} pairs. We first evaluate traditional machine learning classifiers, such as Decision Tree, Random Forest, and XGBoost, and then apply transfer learning to evaluate the performance of the pretrained BERT models. To enhance model interpretability, we apply Explainable AI technique, specifically \gls{lime}, to provide transparency into the decision-making process. Our system achieves accuracy, precision, recall, and F1-score values of 98.66\%, 98.67\%, 98.66\%, and 98.66\%, respectively. This work demonstrates the potential of utilizing transfer learning with BERT-based models for the automated generation of \gls{cam}s, offering high performance and interpretability in educational outcome assessment.}


\maketitle

\section{Introduction}\label{sec:intro}

In higher education, particularly in engineering and technical disciplines, accrediting bodies such as the National Board of Accreditation (NBA) require rigorous mapping between \gls{co}s, \gls{po}s, and \gls{pso}s to ensure academic programs meet predefined educational standards \cite{lavanya2022assessment}. This mapping process, referred to as \gls{cam} alignment, plays a pivotal role in curriculum design, program assessment, and continuous improvement. The \gls{cam} alignment provides a structured approach for determining how individual course outcomes contribute to the achievement of broader program goals. It ensures that each course is aligned with the overall objectives of the program, which is critical for maintaining high academic standards and meeting accreditation requirements.

\gls{co}s define the specific knowledge, skills, and competencies that students are expected to acquire upon completing a course. \gls{po} describe the broader abilities and knowledge that students should possess by the time they complete the entire academic program. \gls{pso} are more specialized, focusing on the particular skills and expertise that students should gain within their specific field of study. Aligning \gls{co}s with \gls{po}s and \gls{pso}s is essential for ensuring that the curriculum is designed to meet both general and discipline-specific educational goals \cite{Admuthe,Mundhe}.

The process of aligning \gls{co}s with \gls{po}s and \gls{pso}s is typically assessed using a scoring system \cite{Yadav}, often referred to as the alignment score, which helps determine the strength of the connection between them. A score of 0 indicates a weak or no alignment, 1 represents a moderate alignment, 2 indicates a strong alignment, and 3 signifies a very strong alignment. Historically, this process has been carried out manually by faculty members. This requires them to evaluate and interpret the relationships between various outcomes. However, the subjective nature of the task, combined with the diversity of terminologies used in \gls{co}, \gls{po}, and \gls{pso} descriptions, makes it time-consuming, prone to inconsistencies, and labor-intensive \cite{Liew}.

Research in this area highlights several challenges in automating \gls{co}-\gls{po}-\gls{pso} alignment, and key questions remain unsolved \cite{Sengupta}:

\begin{itemize}
    \item How can semantic relationships between \gls{co}s, \gls{po}s, and \gls{pso}s be captured effectively to automate the alignment process while minimizing human intervention?
    \item What approaches can address the lack of standardized datasets and terminologies across different institutions and programs?
    \item How can model interpretability be improved to provide educators with insights into automated alignment decisions?
\end{itemize}

In this paper, we present a novel approach that automates the \gls{co} and \gls{po}/\gls{pso} alignment process using pre-trained BERT-based models. This method utilizes transfer learning to fine-tune these models, enabling them to capture the semantic relationships between course and program outcomes. The significance of this work lies in its potential to streamline the construction of course articulation matrices, offering a scalable solution that can be applied to various educational systems and similar alignment tasks in different domains.

The key contributions of this research include:
\begin{itemize}
    \item Preparation of a high-quality, manually curated dataset containing \gls{co}s, \gls{po}, and \gls{pso}, along with their corresponding alignment scores. This dataset forms the foundation for training and evaluating the alignment models, ensuring relevance and quality in the automated process.
    \item Implementation of a synonym-based data augmentation method, where 30\% of the words in \gls{co} and \gls{po}/\gls{pso} descriptions are replaced with semantically similar synonyms. This technique enhances linguistic diversity, addresses class imbalance, and improves the generalization capabilities of the model.
    \item Integration of transfer learning using pre-trained BERT-based models for automating the construction of \gls{cam}. Machine learning classifiers such as Decision Trees, Random Forest, and XGBoost were initially used as benchmarks to evaluate model performance. BERT-based models were then fine-tuned to capture the semantic relationships between \gls{co}s and \gls{po}s or \gls{pso}s. This approach facilitates the extraction of deep semantic features, improving the alignment process and enhancing performance.
    \item Use of \gls{lime} to ensure transparency and interpretability of the alignment process. This enables human-readable explanations, helping educators understand the reasoning behind the model's predictions.
    \item Our approach demonstrates impressive performance on key evaluation metrics, including accuracy, precision, recall, and F1-score. Furthermore, by automating the previously manual process of constructing the \gls{cam}, it saves both time and human resources.

\end{itemize}

The rest of the paper is arranged as follows: Section \ref{sec2} provides a comprehensive overview and analysis of prior research in this domain. Section \ref{sec3} provides an overview of the system methodology, detailing each stage of the process, including the evaluation metrics used to assess model performance. After that, Section \ref{sec4} illustrates the detailed experimental analysis of this study. Finally, Section \ref{sec5} draws the study’s conclusions, along with the plan for further research.

\section{Literature Review}\label{sec2}
In recent years, substantial progress has been made in both curriculum mapping and text classification models, providing significant advancements in educational program evaluation and text analysis tasks. Curriculum mapping has evolved from simple static models to more dynamic and reflective approaches, incorporating innovative frameworks and technological solutions. Text classification, on the other hand, has undergone a revolution with the advent of deep learning models, particularly those built on the Transformer architecture. These models, including BERT and its variants, have drastically improved the ability to understand and classify textual data, making them suitable for tasks like automated alignment of educational objectives.

This literature review will first explore the advancements in curriculum mapping, focusing on various frameworks and methodologies that have emerged to improve the alignment and assessment of educational outcomes. Subsequently, the review will delve into the evolution of text classification models, emphasizing the shift from traditional machine learning approaches to modern deep learning techniques that have reshaped natural language processing tasks. 

The integration of these advancements sets the stage for exploring how they can be applied to enhance automated \gls{cam}, an area that has seen limited exploration of BERT-based models.
\subsection{Advancements in Curriculum Mapping}

Curriculum mapping has evolved into a crucial tool for the evaluation of educational programs, offering a structured approach to aligning instructional practices with intended learning outcomes. Recent studies have significantly advanced the field by introducing reflective practices, innovative technological platforms, and rule-based frameworks that enhance both the alignment and assessment of educational objectives. This section reviews key contributions to the development and application of curriculum mapping techniques.

In their work, Plaza et al. \cite{articlep} highlighted the use of curriculum mapping as a systematic approach to program evaluation and assessment in educational contexts. By employing a descriptive cross-sectional design, their study analyzed existing data from documents detailing student learning outcomes alongside various student and curriculum datasets. The comparative analysis of curriculum maps created by students versus instructors focused on the prioritization of different educational domains, revealing a shared understanding of curricular priorities between students and faculty, and confirming congruence among the intended, delivered, and received curricula.

A different perspective was presented by Spencer et al. \cite{articles}, who proposed an innovative curriculum mapping method centered on reflective practice to assess instructional methods and graduate competencies. Their approach, which encouraged introspective design practices, utilized heat maps to visually represent areas within the curriculum that required development. These maps served as diagnostic tools, identifying gaps and aligning instructional strategies with targeted competencies, ultimately suggesting pathways for curriculum enhancement based on empirical evidence.

Building on the conceptual foundations of curriculum theory in higher education, Linden et al. \cite{inbookl} focused on competency-based and outcome-oriented models. Their study analyzed both the intellectual and historical frameworks underpinning curricular theories and advocated for a unified approach integrating both normative and critical perspectives. The authors emphasized the importance of aligning curricula with higher education standards and recommended routine updates to ensure a balanced and adaptable framework that meets evolving academic and professional needs.

Treadwell et al. \cite{Treadwell2019} introduced a web-based curriculum mapping approach known as the Learning Opportunities, Goals, and Outcome Platform (LOOOP). This innovative tool collected instructors' perspectives on the effectiveness of curriculum design through a survey employing a four-point Likert scale. The findings indicated strong support for LOOOP’s usability and value, suggesting its potential to improve the alignment between curriculum goals and instructional practices while promoting a more structured approach to curriculum mapping.

In a related study, Watson et al. \cite{Watson_Steketee_Mansfield_Moore_Dalziel_Damodaran_Walker_Duvivier_Hu_2020} developed a multidimensional typology for curriculum maps, organizing features into four distinct categories: purpose, product, process, and display. This typology, created through an extensive literature review of higher education, was validated by comparing six curriculum maps from Australian medical schools. Their framework aims to support health education professionals in making informed curriculum mapping decisions, ultimately improving alignment with educational requirements and instructional methodologies.

Lastly, Alshanqiti et al. \cite{articlealsha} proposed an automated, rule-based framework to assess academic curriculum mapping, with a particular focus on the alignment of Course Learning Outcomes (CLOs) with Program Learning Outcomes (PLOs). This framework helps identify misalignments and offers actionable recommendations for enhancing curricula. Collaborating with curriculum specialists, the authors developed a web-based tool to automate the curriculum assessment process using experimental data from real users, though some manual input remains necessary for verifying CLO-to-PLO mappings.

\subsection{Evolution of Text Classification Models}

The task of analyzing and classifying text has a long history, with early efforts relying heavily on handcrafted features and simple machine learning classifiers. One of the most fundamental challenges in text classification is capturing semantic similarity between texts. Early approaches focused on extracting word-level features such as term frequency-inverse document frequency (TF-IDF) and using them in models like \gls{svm} and Naive Bayes for classification tasks \cite{Salton1983}. However, these methods struggled to capture deeper semantic relationships and contextual nuances in text. Researchers began exploring more sophisticated ways to represent text beyond individual words, with approaches such as Latent Semantic Analysis (LSA) \cite{Deerwester1990} and Latent Dirichlet Allocation (LDA) \cite{Blei2003} offering ways to model semantic topics in documents. Despite these advancements, these techniques still lacked the ability to understand the fine-grained relationships between texts, which motivated the exploration of more powerful deep learning models.

\gls{rnn} \cite{Rumelhart1986} and \gls{lstm} \cite{10.1162/neco.1997.9.8.1735} marked a pivotal development in text-to-text analysis. Traditional models like \gls{svm}s and logistic regression were unable to process sequential data effectively. \gls{rnn}s process text data step-by-step, making them ideal for sequence-based tasks such as sentiment analysis and machine translation. However, \gls{rnn}s struggled with long-range dependencies due to the gradient vanishing problem. \gls{lstm}s addressed this by incorporating memory cells and gating mechanisms that allowed them to retain information over long sequences, significantly improving performance on tasks requiring the understanding of long-term context, such as machine translation and speech recognition. While \gls{rnn}s and \gls{lstm}s made significant strides, they still faced challenges in training efficiency and parallelization, which became key considerations for future models.

The introduction of the Transformer model by Vaswani et al. \cite{vaswani2023attentionneed} revolutionized \gls{nlp} by overcoming the limitations of \gls{rnn}s and \gls{lstm}s. The Transformer model introduced the self-attention mechanism, which allowed the model to process all words in a sequence simultaneously, in parallel, rather than sequentially. This parallelization dramatically increased training efficiency and speed, enabling the model to scale effectively to large datasets. Moreover, the self-attention mechanism allowed the Transformer to capture long-range dependencies in text more effectively than \gls{rnn}s and \gls{lstm}s. The Transformer’s ability to process entire input sequences at once eliminated the gradient vanishing problem and enabled the model to learn richer and more comprehensive representations of text. This advancement made Transformers highly effective for a wide range of tasks, including machine translation, document classification, and text generation, setting the stage for the next generation of \gls{nlp} models.

The development of large-scale, pretrained models like BERT (Bidirectional Encoder Representations from Transformers) by Devlin et al. \cite{devlin2019bertpretrainingdeepbidirectional} revolutionized the field of text analysis by introducing transfer learning to \gls{nlp}. BERT was pretrained on vast corpora of text data and fine-tuned for specific tasks, allowing it to learn general language representations that could be adapted to a wide variety of downstream tasks. Unlike previous models, BERT leverages bidirectional context, which enables it to understand the meaning of words in context, making it highly effective for tasks like question answering, named entity recognition, and sentiment analysis. Following BERT’s success, several variants were introduced, including DistilBERT \cite{sanh2020distilbertdistilledversionbert}, a smaller and faster version of BERT, ALBERT \cite{lan2020albertlitebertselfsupervised}, which reduces model size through parameter sharing, and RoBERTa \cite{liu2019robertarobustlyoptimizedbert}, which optimizes the training process by removing certain training constraints. These models built upon BERT’s architecture and further advanced the state of the art in a range of \gls{nlp} tasks.

While previous studies have made significant progress in advancing automated \gls{cam}, most have not utilized BERT-based models. The lack of large, diverse datasets remains a key limitation; however, transfer learning presents a promising solution by fine-tuning pre-trained models on smaller, task-specific datasets. By incorporating BERT family models, we aim to enhance the accuracy and generalization of \gls{cam}, ultimately improving performance even with limited data.

\section{Methodology}\label{sec3}
The workflow of our project follows a structured approach to automate the generation of Course Articulation Matrices. Initially, we collect a dataset containing \gls{co}s and \gls{po}s/\gls{pso}s. Data augmentation techniques are then applied to expand the dataset, ensuring model robustness. Next, we train various models from the BERT family to capture the nuanced relationships between \gls{co}s and \gls{po}s/\gls{pso}s. Model performance is evaluated using standard metrics to validate effectiveness. To gain insights into model decisions, we conduct a misclassification analysis and apply \gls{lime}, which provides transparency in the model's decision-making process. This workflow enables a systematic, explainable approach to \gls{cam} generation, enhancing alignment accuracy and interpretability.
\begin{figure}[ht]
    \centering
    \includegraphics[width=\textwidth]{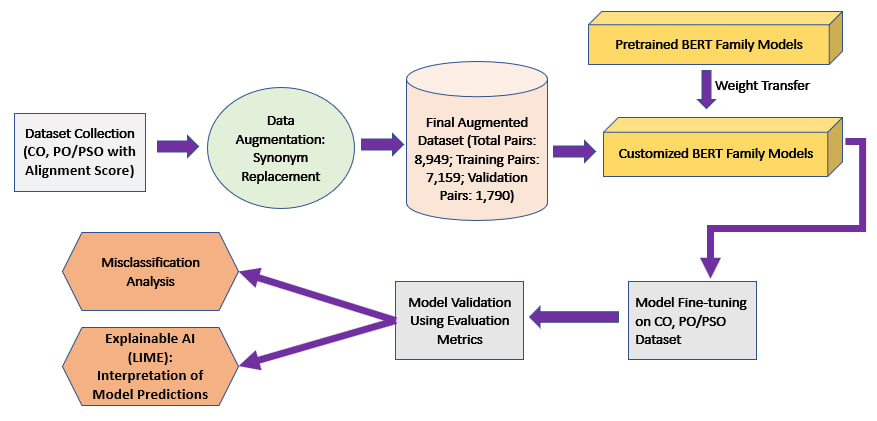}
    \caption{Workflow of the proposed model using transfer learning, from data collection and augmentation to BERT fine-tuning and model interpretation.}
    \label{fig:td}
\end{figure}
\subsection{Data Collection and Preprocessing}

\subsubsection{Dataset Preparation}
The dataset used in this study was collected from 22 courses offered by \href{https://cgu-odisha.ac.in/}{C. V. Raman Global University}, India, and comprises 1,840 pairs of \gls{co} and \gls{po} or \gls{pso}, each paired with pre-assigned alignment scores. These scores, which range from 0 (no alignment) to 3 (full alignment), were sourced from the university's pre-curated \gls{cam}. Both the original and augmented versions of the dataset are available at \href{https://github.com/natenaile/BERT-Based-Approach-for-Automating-Course-Articulation-Matrix-Construction-with-Explainable-AI/tree/main/DATASET}{GitHub}. While the original dataset established a structural foundation, only the augmented version was utilized for model training and evaluation.

\subsubsection{Data Augmentation}
To address class imbalance and enhance model generalization, we applied data augmentation to the \gls{co} and \gls{po}/\gls{pso} descriptions. Specifically, 30\% of words in each description were randomly replaced with semantically similar synonyms, maintaining the original meaning while introducing linguistic variety.

Augmentation was applied across all alignment score classes (0–3). The alignment score distribution after augmentation included 2500 pairs for Score 0, 2158 pairs for Score 1, 2140 pairs for Score 2, and 2151 pairs for Score 3, resulting in a balanced dataset with a total of 8949 pairs. The dataset was then shuffled to avoid ordering bias, ensuring improved generalization during model training and validation.
\subsubsection{Data Preprocessing}
Textual data preprocessing was performed using the tokenization and preprocessing capabilities of the BERT family of models (BERT Base, DistilBERT, ALBERT, RoBERTa). Each model's pre-trained tokenizer was employed to handle key preprocessing tasks, including tokenization, conversion to input IDs, and generation of attention masks. This automated process ensures efficient handling of variable-length inputs while maintaining compatibility with the downstream tasks. By leveraging the built-in preprocessing functions of these models, we streamlined the pipeline and eliminated the need for manual intervention, thereby ensuring consistent and high-quality input data for model training and evaluation.

\subsection{Model Description}

\subsubsection{Machine Learning Classifiers}
As a baseline for comparison, three traditional classifiers: Decision Tree, Random Forest, and XGBoost, were initially evaluated for classifying the alignment between \gls{co} and \gls{po} or \gls{pso}. These models were selected to provide a benchmark for the task, assessing their ability to handle complex relationships between \gls{co} and \gls{po}/\gls{pso} descriptions. Although they were not the primary focus of this study, their performance served as a point of reference for evaluating the more advanced transformer models. The characteristics and advantages of each classifier are summarized in Table \ref{tab:tree_comparisons}.
\begin{table*}[htbp] 
\centering
\caption{Comparison of Selected Machine Learning Classifiers for Automating \gls{cam} Construction.}
\label{tab:tree_comparisons}
\small 
\begin{tabularx}{\textwidth}{lXXXXXX}
\toprule
\textbf{Model} & \textbf{Author(s)} & \textbf{Description} & \textbf{Key Features} \\ 
\midrule
Decision Tree   & Breiman et al. \cite{Breiman1986}       & A decision tree model that splits data into branches, making decisions based on feature values. & Simple and interpretable; prone to overfitting with complex datasets. \\ 
\midrule
Random Forest   & Breiman \cite{Breiman2001}       & An ensemble of decision trees that combines their predictions for more robust performance. & Reduces overfitting compared to single decision trees; effective for complex tasks and imbalanced datasets. \\ 
\midrule
XGBoost         & Chen and Guestrin \cite{Chen2016} & A gradient boosting method that builds trees sequentially to correct previous errors. & High accuracy, scalability, and speed; includes regularization to prevent overfitting. \\ 
\bottomrule
\end{tabularx}
\end{table*}

\subsubsection{Pre-trained Transformer Models}
The pre-trained transformer models used in this study include BERT Base, DistilBERT, ALBERT, and RoBERTa, each offering unique advantages for the alignment task between \gls{co} and \gls{po} or \gls{pso}. The selection of these models is justified by their strong performance in natural language processing tasks, particularly in understanding contextual relationships between paired textual inputs. Utilizing transfer learning enhances computational efficiency, as these models have been pre-trained on large text corpora, leading to improved accuracy in predicting alignment scores. Table \ref{tab:transformers} highlights the architectural characteristics that make each model particularly suitable for this task.

\begin{table*}[htbp] 
\centering
\caption{Comparison of selected Transformer Models for automating \gls{cam} construction.}
\label{tab:transformers}
\small 
\begin{tabularx}{\textwidth}{lXXXXXX}
\toprule
\textbf{Model} & \textbf{Author(s)} & \textbf{Layer Details} & \textbf{Key Features} \\ 
\midrule
BERT Base       & Devlin et al. \cite{devlin2019bertpretrainingdeepbidirectional} & 12 Transformer layers, 768 hidden units, 12 attention heads, 110M parameters & Pretrained using masked language modeling (MLM). Introduces Next Sentence Prediction (NSP) task for training. \\ 
\midrule
DistilBERT Base & Sanh et al. \cite{sanh2020distilbertdistilledversionbert}   & 6 Transformer layers, 768 hidden units, 12 attention heads, 66M parameters  & A smaller, faster, and lighter version of BERT, achieved via knowledge distillation. Retains 97\% of BERT's performance. \\ 
\midrule
ALBERT Base     & Lan et al. \cite{lan2020albertlitebertselfsupervised}    & 12 Transformer layers, 768 hidden units, 12 attention heads, 12M parameters (shared weights) & Parameter reduction via weight sharing across layers and factorized embedding parameters. Reduces the model size drastically. \\ 
\midrule
RoBERTa Base    & Liu et al. \cite{liu2019robertarobustlyoptimizedbert}    & 12 Transformer layers, 768 hidden units, 12 attention heads, 125M parameters & Trained on more data, removes the NSP task, and uses more training steps. Employs dynamic masking and larger batch sizes for better performance. \\ 
\bottomrule
\end{tabularx}
\end{table*}

\subsubsection{Transfer Learning}
Pan and Yang \cite{panarticle} discussed transfer learning as a powerful technique in deep learning that allows a model trained on one task to be effectively applied to another, often related task. It is particularly useful when labeled data is scarce, as it leverages the knowledge gained from large datasets. There are generally two approaches to transfer learning:

\begin{itemize}
    \item Weight Initialization: In this method, the pre-trained model’s weights are used as the initial values for the new task, speeding up convergence and enhancing performance. The entire model is then trained on the new dataset, fine-tuning the weights to adapt to task-specific patterns.
    
    \item Feature Extraction: This approach involves freezing some layers of the pre-trained model to retain learned features while fine-tuning other layers for the new task, improving performance by adapting to the specifics of the new dataset.
\end{itemize}

In this study, we adopted the first approach (weight initialization) because it allows the model to fully learn the intricate relationships between \gls{co} and \gls{po} or \gls{pso} by fine-tuning all layers on the augmented dataset of \gls{co} and \gls{po}/\gls{pso} pairs.

\subsection{Model optimization and training}

\subsubsection{Adam Optimizer}

Kingma and Ba \cite{kingma2017adammethodstochasticoptimization} introduced a widely used optimization algorithm in deep learning that efficiently adjusts the learning rates of model parameters. We employed this optimizer due to its ability to combine the benefits of two other optimizers, AdaGrad and RMSProp, for faster convergence. It computes adaptive learning rates for each parameter by considering both the first moment (mean) and the second moment (uncentered variance) of the gradients. The update rules for the Adam optimizer are shown in the respective formulas \ref{eq:firstmom}-\ref{eq:pareq}.

First moment estimate: Updates an exponentially weighted average of gradients, which represents the mean direction of the gradient over time.
\begin{equation}
m_t = \beta_1 m_{t-1} + (1 - \beta_1) g_t
\label{eq:firstmom}
\end{equation}

Second moment estimate: Updates an exponentially weighted average of the squared gradients, capturing the variance or magnitude of the gradient.
\begin{equation}
v_t = \beta_2 v_{t-1} + (1 - \beta_2) g_t^2
\label{eq:secomom}
\end{equation}

Bias correction for the first moment: Adjusts the first moment estimate to account for its bias, especially during the initial iterations when it is close to zero.
\begin{equation}
\hat{m}_t = \frac{m_t}{1 - \beta_1^t}
\label{eq:biasfirst}
\end{equation}

Bias correction for the second moment: Adjusts the second moment estimate to account for its bias, particularly in early iterations when it is influenced by initialization.
\begin{equation}
\hat{v}_t = \frac{v_t}{1 - \beta_2^t}
\label{eq:biassecond}
\end{equation}

Parameter update: Uses the corrected first and second moment estimates to update the model parameters, incorporating the learning rate for stability.
\begin{equation}
\theta_t = \theta_{t-1} - \frac{\alpha \hat{m}_t}{\sqrt{\hat{v}_t} + \epsilon}
\label{eq:pareq}
\end{equation}

In these equations:
\begin{itemize}
    \item \( g_t \) represents the gradient of the loss function with respect to the parameters.
    \item \( \beta_1 \) and \( \beta_2 \) are hyperparameters that determine the decay rates of the moving averages (typically set to \( 0.9 \) and \( 0.999 \), respectively).
    \item \( \alpha \) is the learning rate, initialized in this implementation as \( 5 \times 10^{-5} \).
    \item \( \epsilon \) is a small constant added to prevent division by zero.
\end{itemize}

The Adam optimizer is particularly effective for training deep learning models due to its ability to adaptively adjust the learning rates based on the dynamics of the gradients.

\subsubsection{Categorical Loss Function}

We utilized this loss function, which was introduced by Goodfellow et al \cite{Goodfellow-et-al-2016}, for its effectiveness in multi-class classification tasks. This function is well-suited for predicting discrete class labels, where each label corresponds to a score (0, 1, 2, 3) representing the degree of alignment for \gls{co}-\gls{po}/\gls{pso} pairs.

The mathematical formulation for categorical loss in this context is given by Formula \ref{eq:clf}.

\begin{equation}
L(y, \hat{y}) = -\frac{1}{N} \sum_{i=1}^{N} \sum_{j=1}^{C} y_{ij} \log(\hat{y}_{ij})
\label{eq:clf}
\end{equation}

Where:
\begin{itemize}
    \item \(N\) is the number of samples, where each sample corresponds to a pair of \gls{co} and \gls{po}/\gls{pso}.
    \item \(C\) is the number of classes, representing the possible alignment scores (0, 1, 2, or 3).
    \item \(y_{ij}\) is the true label for sample \(i\) and class \(j\), where \(y_{ij}\) takes a value of 0, 1, 2, or 3, indicating the true alignment score for the corresponding \gls{co}-\gls{po}/\gls{pso} pair.
    \item \(\hat{y}_{ij}\) is the predicted probability for class \(j\) (alignment score) for sample \(i\).
\end{itemize}

The goal is to improve the model's accuracy in predicting alignment scores for \gls{co}-\gls{po}/\gls{pso} pairs. Minimizing the cross-entropy ensures the model's predicted probabilities (\(\hat{y}_{ij}\)) closely match the true alignment labels (\(y_{ij}\)), enhancing performance in the automated generation of the \gls{cam}.

\subsubsection{Model Hyperparameters and Training}
The dataset, consisting of 8,949 pairs, was split into training (80\% or 7,159 pairs) and validation (20\% or 1,790 pairs) sets. Both machine learning and transformer-based models were trained on the training set, while the validation set was used to evaluate their performance.

Traditional machine learning models were trained with selected hyperparameters to predict the four alignment scores in the \gls{cam}. The Decision Tree classifier was configured with the Gini impurity criterion, a maximum depth of 10, a minimum of 5 samples required to split an internal node, and 3 samples required at each leaf node. The Random Forest classifier utilized 200 estimators, a maximum depth of 30, and a minimum of 5 samples to split, with 3 samples per leaf. The number of features considered for splitting was set to the square root of the total number of features, and bootstrap sampling was enabled. The XGBoost model was defined with 200 estimators, a learning rate of 0.1, and additional parameters tuned for optimal performance.

The pre-trained models, specifically from the BERT family (DistilBERT, BERT Base, ALBERT, and RoBERTa), were adapted for the task of automated \gls{cam} generation by modifying the last fully connected layer to classify four alignment score classes (0, 1, 2, 3). All transformer layers were unfrozen, allowing the pre-trained weights to adjust to the specific patterns in the alignment of \gls{co}s and \gls{po}s/\gls{pso}s. The models were trained with an initial learning rate of 5e-5 and a batch size of 64 over 10 epochs, using the AdamW optimizer. The loss function used for training was cross-entropy, which is implicitly handled by the models for classification tasks. The maximum sequence length was set to 512 to accommodate the input data. 

The choice of hyperparameters was determined through experimentation and fine-tuning.
\subsection{Evaluation Metrics}

In this section, we describe the evaluation metrics used to assess model performance in the \gls{cam} automation task for alignment score prediction, along with their respective formulas \ref{eq:accuracy}-\ref{eq:f1score}. These metrics: accuracy, precision, recall, and F1-score are essential to understanding the strengths and weaknesses of the models.

The following terms are common to all metrics:
\begin{itemize}
    \item \(TP\) = True Positives (correctly predicted positive instances)
    \item \(TN\) = True Negatives (correctly predicted negative instances)
    \item \(FP\) = False Positives (incorrectly predicted positive instances)
    \item \(FN\) = False Negatives (incorrectly predicted negative instances)
\end{itemize}

Accuracy is a fundamental metric used to evaluate the performance of a classification model, defined as the ratio of correctly predicted instances to the total instances in the dataset. The formula for accuracy is given by:

\begin{equation}
\text{Accuracy} = \frac{TP + TN}{TP + TN + FP + FN} \label{eq:accuracy}
\end{equation}
 
Precision measures the proportion of true positive predictions among all positive predictions. It is calculated as follows:

\begin{equation}
\text{Precision} = \frac{TP}{TP + FP} \label{eq:precision}
\end{equation}

Recall (also known as sensitivity) assesses the proportion of true positive predictions relative to the actual positives. It is calculated using the formula:

\begin{equation}
\text{Recall} = \frac{TP}{TP + FN} \label{eq:recall}
\end{equation}
 
The F1-score is the harmonic mean of precision and recall, providing a balance between the two metrics. It is calculated as follows:

\begin{equation}
\text{F1-score} = 2 \times \frac{\text{Precision} \times \text{Recall}}{\text{Precision} + \text{Recall}} \label{eq:f1score}
\end{equation}

\section{Experiments}\label{sec4}
This section presents an analysis of the experimental results, evaluating the effectiveness of our proposed approach for automating the \gls{cam} generation. Table \ref{param} summarizes the parameters used in our experiments to facilitate reproducibility of results.
\begin{table*}[htbp] 
\centering
\caption{Parameters Used in the Experiment.}
\label{param}
\small 
\begin{tabularx}{\textwidth}{lXXXXXX}
\toprule
\textbf{Parameter Name} & \textbf{Values} \\ 
\midrule
Hyperparameters         & Batch Size: 64, Learning Rate: 5e-5, Epochs: 10, Max Sequence Length: 512 \\ 
\midrule
System Specifications   & CPU: Intel i9, RAM: 16 GB, GPU: NVIDIA RTX A2000, Storage: 1 TB HDD \\ 
\midrule
Operating System        & Windows 11 \\ 
\midrule
Implementation          & PyTorch and Hugging Face Transformers \\ 
\midrule
Dataset                 & Total Pairs: 8,949; Training Pairs: 7,159; Validation Pairs: 1,790 \\ 
\midrule
Model Types             & BERT-base, DistilBERT, ALBERT, RoBERTa \\ 
\midrule
Task                    & Predicting four alignment scores (0, 1, 2, 3) for CO-PO/PSO pairs \\ 
\bottomrule
\end{tabularx}
\end{table*}

\subsection{Result analysis and discussion}
Initially, machine learning classifiers, including Decision Tree, Random Forest, and XGBoost, were evaluated for predicting the four alignment scores (0, 1, 2, 3) for \gls{co}-\gls{po}/\gls{pso} pairs, with the highest accuracy of 87.93\% achieved by Random Forest and the lowest accuracy of 79.11\% by Decision Tree, as summarized in Table \ref{tab:ml_performance}.

Subsequently, the four transformer-based models: BERT-base, DistilBERT, ALBERT, and RoBERTa, were evaluated for automating the construction of the \gls{cam}. Among these, DistilBERT emerged as the top performer, achieving an accuracy of 98.66\%, precision of 98.67\%, recall of 98.66\%, and an F1-score of 98.66\%. Its performance can be attributed to its streamlined architecture, which maintains high accuracy while offering computational efficiency. In contrast, ALBERT was the lowest performer, with an accuracy of 96.87\%, precision of 96.86\%, recall of 96.87\%, and an F1-score of 96.86\%. Despite being designed for efficiency with parameter sharing, ALBERT's ability to capture the complexities of the task was limited, leading to suboptimal results. BERT-base and RoBERTa performed similarly, with BERT-base achieving an accuracy of 98.32\% and RoBERTa slightly lower at 97.99\%. The differences in their performance can be attributed to variations in their training strategies, with RoBERTa utilizing a larger training corpus and extended training duration, yet still slightly underperforming compared to BERT-base.

In terms of training time, DistilBERT was the most efficient, requiring only 523.69 seconds to complete training, highlighting its balance of performance and computational efficiency. Conversely, ALBERT had the longest training time at 2077.65 seconds. Although designed to be parameter-efficient, ALBERT’s intricate architecture and optimization process contributed to its extended training time.

In summary, DistilBERT stands out as the most effective and time-efficient model for the task, offering an optimal combination of high performance and rapid training. These results emphasize the importance of selecting models that balance accuracy and training efficiency. The detailed results are summarized in Table \ref{tab:bert performance}.
\subsection{Comparing Machine Learning and Transformer-Based Models for Automated \gls{cam} Construction}
Our results in Table \ref{tab:ml_performance} and Table \ref{tab:bert performance} show that machine learning models like Decision Tree, Random Forest, and XGBoost are effective for structured data but struggle with capturing complex patterns in textual data without extensive feature engineering. In contrast, transformer models such as BERT and its variants (DistilBERT, ALBERT, RoBERTa) excel in understanding context and relationships within text, leveraging attention mechanisms for better performance. While transformer models deliver superior accuracy, they often require significantly more computational resources and longer training times. The decision between machine learning and transformers depends on finding a balance between accuracy and computational efficiency. Overall, transformer models like DistilBERT provide an ideal combination of both high performance and efficiency.
\begin{table*}[htbp] 
\centering
\caption{Performance Metrics of Machine Learning Models for Automating \gls{cam} Construction on the Validation Dataset.}
\label{tab:ml_performance}
\small 
\begin{tabularx}{\textwidth}{lXXXXXX}
\toprule
\textbf{Model}         & \textbf{Accuracy (\%)} & \textbf{Precision (\%)} & \textbf{Recall (\%)} & \textbf{F1-score (\%)} \\ 
\midrule
Decision Tree          & 79.11                 & 79.46                   & 79.11                & 78.98                 \\ 
\midrule
Random Forest          & \underline{\textbf{87.93}}                 & \underline{\textbf{87.89}}                   & \underline{\textbf{87.93}}                & \underline{\textbf{87.90}}                 \\ 
\midrule
XGBoost                & 86.42                 & 86.42                   & 86.42                & 86.37                 \\ 
\bottomrule
\end{tabularx}
\end{table*}

\begin{table*}[htbp] 
\centering
\caption{Performance Metrics of Transformer Models for Automating \gls{cam} Construction on the Validation Dataset.}
\label{tab:bert performance}
\small 
\begin{tabularx}{\textwidth}{lXXXXXX}
\toprule
\textbf{Model}         & \textbf{Accuracy (\%)} & \textbf{Precision (\%)} & \textbf{Recall (\%)} & \textbf{F1-score (\%)}  \\ 
\midrule
BERT-base              & 98.32                 & 98.33                   & 98.32                & 98.32                                        \\ 
\midrule
DistilBERT             & \underline{\textbf{98.66}}                 & \underline{\textbf{98.67}}                   & \underline{\textbf{98.66}}                & \underline{\textbf{98.66}}                     \\ 
\midrule
ALBERT                 & 96.87                 & 96.86                   & 96.87                & 96.86                                       \\ 
\midrule
RoBERTa                & 97.99                 & 97.99                   & 97.99                & 97.99                                       \\ 
\bottomrule
\end{tabularx}
\end{table*}

\subsection{Results Visualization}

\begin{figure}[ht]
    \centering
    \includegraphics[width=\textwidth]{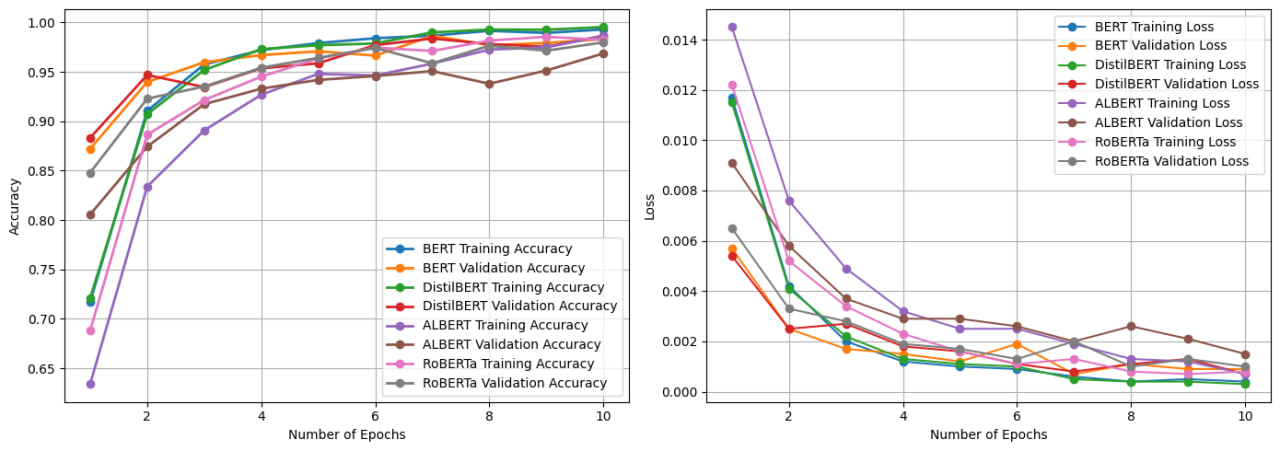}
    \caption{Visualization of Accuracy and Loss Graphs with Respect to Epochs for the \gls{cam} Construction Task.}
    \label{fig:loss acc}
\end{figure}

\begin{figure}[ht]
    \centering
    \includegraphics[width=\textwidth]{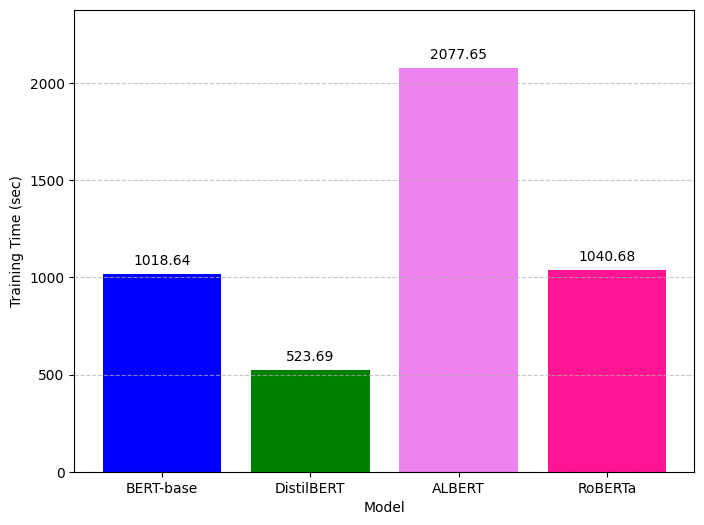}
    \caption{Comparison of training times for BERT-base, DistilBERT, ALBERT, and RoBERTa with colored bars indicating each model.}
    \label{fig:time}
\end{figure}

The performance of the transformer-based models is evaluated through two key metrics: loss and accuracy, presented in two plots that illustrate the training and validation results across 10 epochs for all four models, as shown in Figure~\ref{fig:loss acc}. The comparison of the training times for BERT-base, DistilBERT, ALBERT, and RoBERTa is illustrated in Figure \ref{fig:time}.

\begin{figure}[ht]
    \centering
    \includegraphics[width=\textwidth]{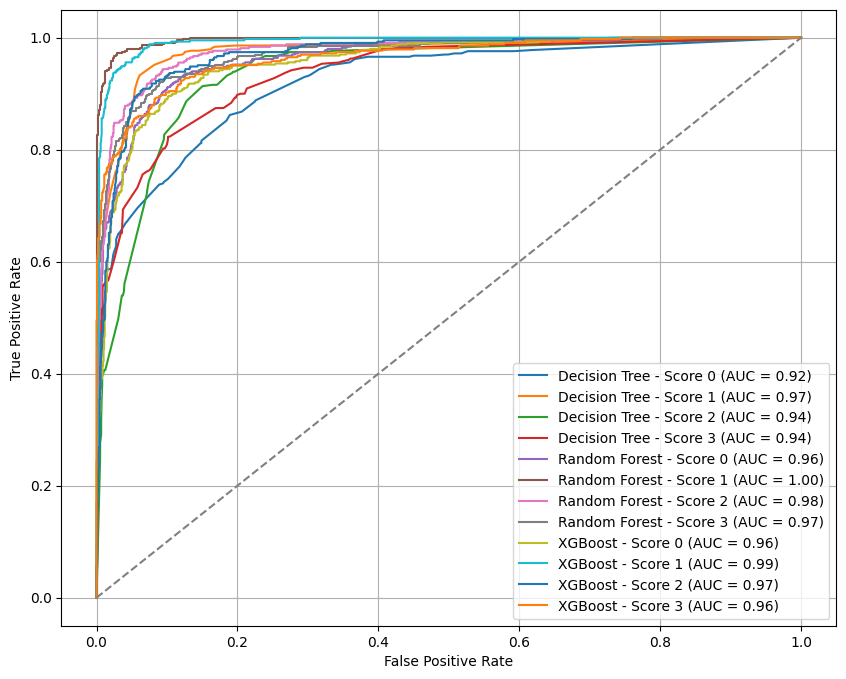}
    \caption{Visualization of ROC Curves for Machine Learning Classifiers in the \gls{cam} Construction Task.}
    \label{fig:roc}
\end{figure}

The ROC curves for the machine learning classifiers are presented in Figure \ref{fig:roc}, highlighting the performance of each classifiers in distinguishing between the different classes.
\subsubsection{ROC Curve}
The performance of the machine learning classifiers, specifically the Decision Tree, Random Forest, and XGBoost, is evaluated using ROC curves across the four score classes, as shown in Figure \ref{fig:roc}. The AUC values for all models are consistently above 0.90, indicating strong discriminatory performance in identifying each class. Despite these high AUC scores, the models show relatively lower validation accuracy, precision, and F1 scores, which suggests that while the models are effective at distinguishing between classes, they may struggle with correctly assigning the final class label.

\subsubsection{Loss Curve}

The left plot in Figure~\ref{fig:loss acc} depicts the loss curve, illustrating the progression of training and validation loss over the epochs. A reduction in loss values indicates effective learning. All four transformer-based models consistently exhibit a decline in both training and validation loss, reflecting enhanced performance at each epoch. Notably, none of the models demonstrate overfitting, as evidenced by stable loss values. The gap between the training and validation loss lines serves as a measure of the models' generalization to unseen data. A smaller gap signifies better generalization, whereas a larger gap may suggest potential overfitting.

\subsubsection{Accuracy Curve}

The right plot in Figure~\ref{fig:loss acc} shows the accuracy curve, which illustrates the training and validation accuracy across the epochs. An upward trend in this curve indicates improvements in classification performance. The plot demonstrates a consistent rise in accuracy for each of the four transformer-based models, suggesting robust model performance. The closeness of the training and validation accuracy lines further indicates that the models are free from overfitting, as they maintain similar accuracy levels throughout the training process.

Overall, these visualizations elucidate the training dynamics of the models, demonstrating their capacity to learn from the data effectively.
\subsection{Effect of Data Augmentation on Model Performance.}
The impact of data augmentation on model performance was evaluated using DistilBERT, the best-performing model, under three experimental settings, as detailed in Table \ref{tab:aug}. Without augmentation, the model achieved a baseline accuracy of 83.70\%. Incorporating data augmentation without shuffling significantly improved accuracy to 97.60\%. The highest performance, with an accuracy of 98.66\%, was attained when data augmentation was combined with shuffling, underscoring the importance of mitigating order bias to optimize model performance.
\begin{table*}[htbp] 
\centering
\caption{Performance of DistilBERT for Automating \gls{cam} Construction on the Validation Dataset.}
\label{tab:aug}
\small 
\begin{tabularx}{\textwidth}{lXXXXXX}
\toprule
\textbf{Model}         & \textbf{Accuracy (\%)} & \textbf{Precision (\%)} & \textbf{Recall (\%)} & \textbf{F1-score (\%)} \\ 
\midrule
Pre-Augmentation          & 83.70                 & 82.90                  & 83.70                & 82.18                \\ 
\midrule
Augmentation (without shuffling)  & 97.60                 & 97.59                   & 97.60                & 97.59                 \\ 
\midrule
Augmentation (with shuffling) & \underline{\textbf{98.66}}                 & \underline{\textbf{98.67}}                   & \underline{\textbf{98.66}}                & \underline{\textbf{98.66}}                     \\ 
\bottomrule
\end{tabularx}
\end{table*}

\subsection{Misclassification Analysis}

Misclassification analysis is essential for understanding the limitations of the model and identifying which alignment scores are most frequently confused. This analysis provides insights into patterns in errors, enabling targeted improvements for future research and model refinements. The analysis was performed using our best-performing model, DistilBERT. Out of the total validation dataset of 1790 \gls{co}-\gls{po}/\gls{pso} pairs, the model produced 24 misclassifications.
\begin{figure}[ht]
    \centering
    \includegraphics[width=\textwidth]{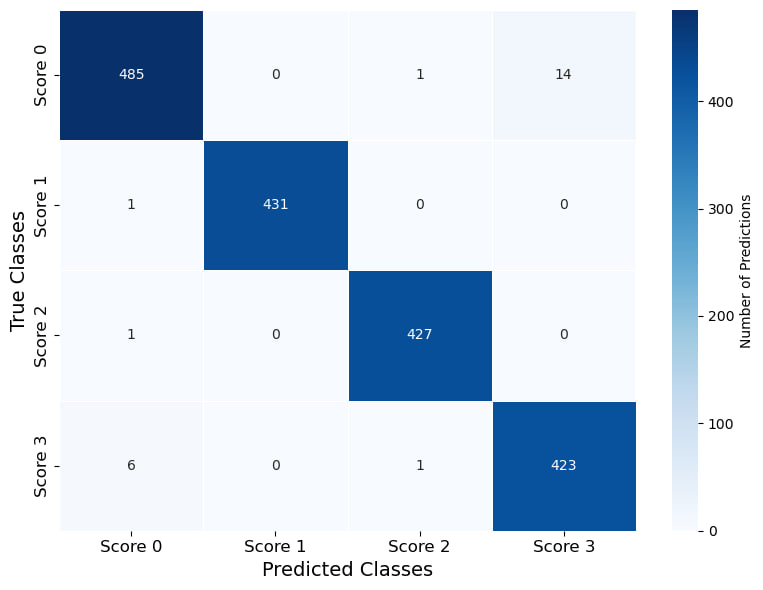}
    \caption{Confusion Matrix for \gls{co}-\gls{po}/\gls{pso} Alignment Score Prediction on the Validation Dataset.}
    \label{fig:conf}
\end{figure}
\begin{figure}[ht]
    \centering
    \includegraphics[width=\textwidth]{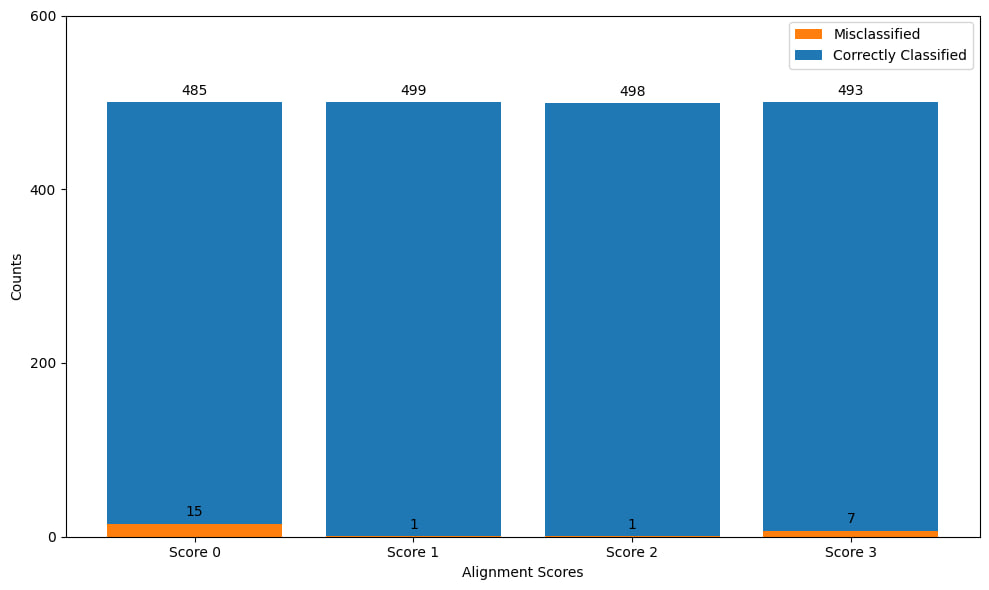}
    \caption{Visualization of Correctly Classified and Misclassified Instance Counts for Each Alignment Score on Validation Dataset.}
    \label{fig:bar}
\end{figure}

\subsubsection{Misclassified Alignment Scores}

The CO-PO/PSO alignment process involves assigning alignment scores (0, 1, 2, 3) to indicate the degree of alignment between each \gls{co} and \gls{po} or \gls{pso}. Due to subtle differences in learning objectives, certain alignment scores are difficult to distinguish. Table \ref{tab:misclassification} lists the misclassified alignment scores along with their respective misclassification rates. For example, alignment scores 0 (weak alignment) and 3 (strong alignment) are frequently confused due to their conceptual overlap. Among 500 instances of score 0, 14 were misclassified as score 3.

The Misclassification Rate for Score A misclassified as Score B is calculated as follows:

\begin{equation}
MR(\text{Score A} \rightarrow \text{Score B}) = \frac{N_{\text{misclass}}}{N_{\text{total}}} \label{eq:mr}
\end{equation}

In this equation:
\begin{itemize}
    \item \( MR \) represents the Misclassification Rate,
    \item \( \text{Score A} \) is the actual alignment score,
    \item \( \text{Score B} \) is the misclassified alignment score,
    \item \( N_{\text{misclass}} \) represents the number of instances of Score A misclassified as Score B,
    \item \( N_{\text{total}} \) is the total number of instances of Score A.
\end{itemize}
\begin{table*}[htbp]
\centering
\caption{Misclassification Analysis for Commonly Confused Alignment Scores on Validation Dataset.}
\label{tab:misclassification}
\small
\begin{tabularx}{\textwidth}{lXXX}
\toprule
\textbf{Alignment Score Pair} & \textbf{Instances (Misclassified / Total)} & \textbf{Misclassification Rate (\%)} \\
\midrule
Score 0 $\rightarrow$ Score 3 & 14 / 500 & 2.800 \\
\midrule
Score 0 $\rightarrow$ Score 2 & 1 / 500  & 0.200 \\
\midrule
Score 3 $\rightarrow$ Score 0 & 6 / 430  & 1.395 \\
\midrule
Score 3 $\rightarrow$ Score 2 & 1 / 430  & 0.233 \\
\midrule
Score 1 $\rightarrow$ Score 0 & 1 / 432  & 0.231 \\
\midrule
Score 2 $\rightarrow$ Score 0 & 1 / 428  & 0.234 \\
\bottomrule
\end{tabularx}
\end{table*}

\subsubsection{Visual Representation of Misclassified Samples Analysis.}
To further analyze misclassifications, a confusion matrix was generated for the validation set, as shown in Figure \ref{fig:conf}. The confusion matrix highlights the alignment score pairs with the highest misclassification rates, providing a visual representation of where the model confuses one alignment score for another. Figure \ref{fig:bar} presents the number of correctly classified and misclassified instances for each alignment score.

\begin{figure}[ht]
    \centering
    \includegraphics[width=\textwidth]{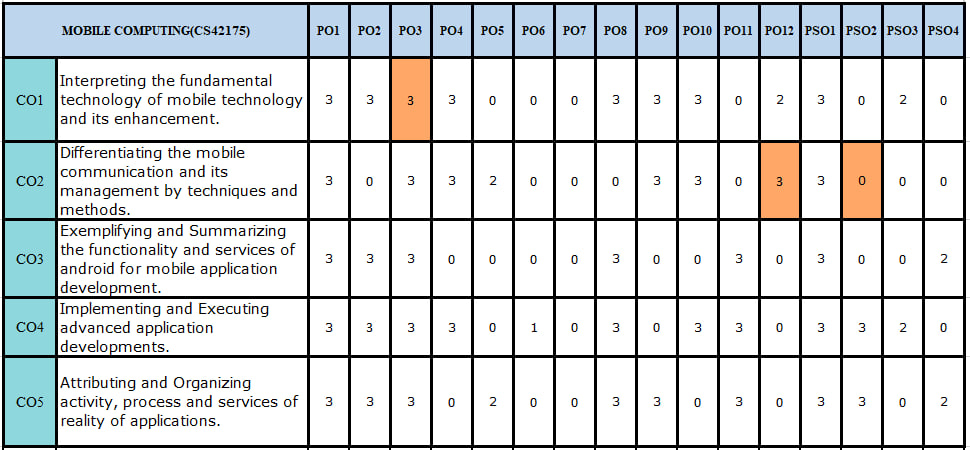}
    \caption{Automated \gls{cam} for the most misclassified course, with misclassified alignment score cells highlighted in orange.}
    \label{fig:cam}
\end{figure}

We also present a sample course from the Automated \gls{cam}, which exhibits a higher number of misclassified cells compared to other courses in our validation set, as shown in Figure \ref{fig:cam}. The orange cells denote the three misclassified instances.

\begin{figure}[ht]
    \centering
    \includegraphics[width=\textwidth]{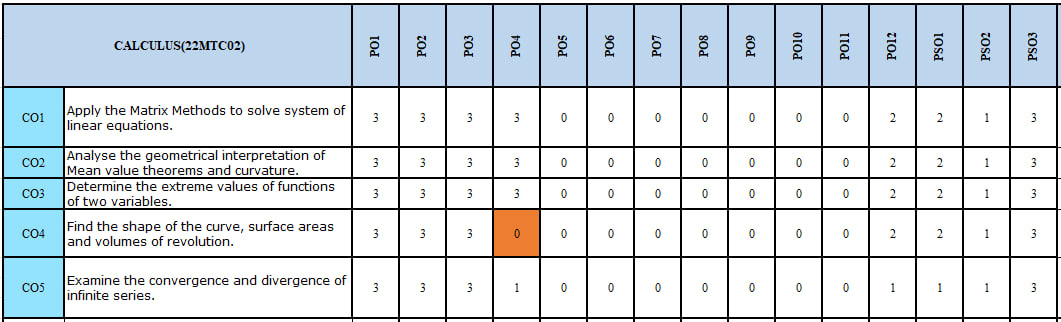}
    \caption{Automated \gls{cam} for a randomly selected course from \gls{cbit}, India, showing one misclassified cell highlighted in orange.}
    \label{fig:cam1}
\end{figure}

Additionally, Figure \ref{fig:cam1} presents the automated CAM for a randomly selected course from \gls{cbit}, India, used as a test case for our model. In this case, only one misclassified cell is observed, highlighted in orange. This analysis demonstrates that our model performs well even when applied to data from an institution not included in the training and validation sets. It suggests that the model has strong generalization capabilities, indicating its potential for large-scale applications across diverse educational institutions.
\subsection{Interpretability of \gls{cam} Automation}
\begin{figure}[ht]
    \centering
    \includegraphics[width=\textwidth]{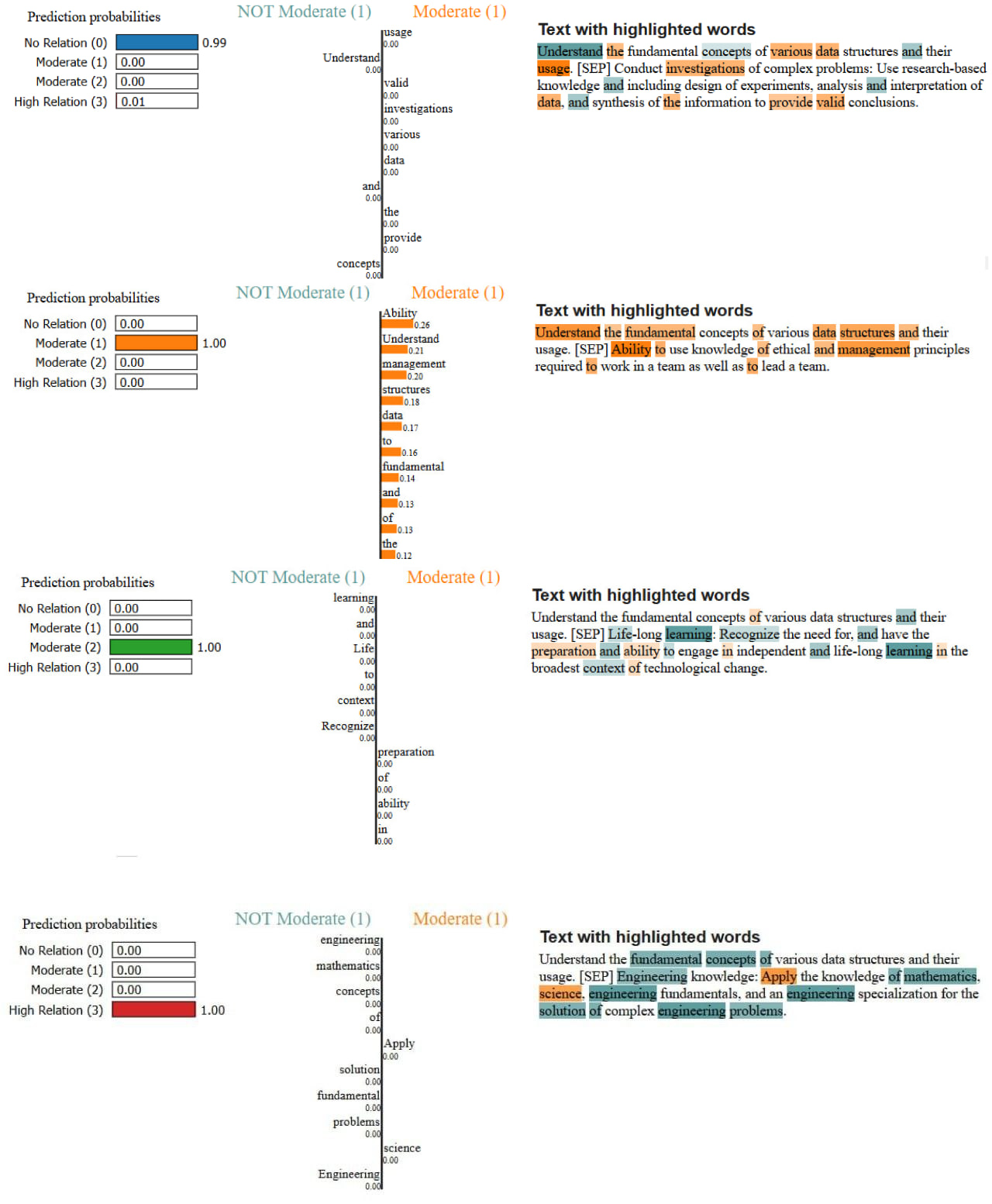}
    \caption{LIME Explanations for CO-PO/PSO Alignment Score Predictions.}
    \label{fig:lime}
\end{figure}

To ensure transparency in the automated \gls{cam} alignment process and validate the model's predictions, we utilized \gls{lime}, introduced by Ribeiro et al. \cite{ribeiro2016why}, as an explainability tool. \gls{lime} enables a localized understanding of the model’s prediction for each \gls{co} and \gls{po}/\gls{pso} alignment pair, providing insight into the predicted alignment scores. We conducted experiments on four distinct \gls{co}-\gls{po} alignment scenarios to examine the model's interpretability and predictive accuracy across varying levels of alignment strength, as shown in Figure \ref{fig:lime}.

\textbf{Non-Relation Case (Alignment Score 0):} In the first test, the \gls{co} was aligned with a \gls{po}, resulting in an alignment score of 0, which indicates no meaningful semantic relationship between the two. \gls{lime} displayed a prediction probability of 0.99 for score 0 (non-relation) and only 0.01 for score 3 (strong relation). In this case, certain words were highlighted in orange, showing that these words had a moderate influence but were not strong enough to shift the alignment to a higher score. This is shown in the first plot of Figure \ref{fig:lime}.

\textbf{Moderate Relation Case (Alignment Score 1):} In the second test, the same \gls{co} aligned with a different \gls{po}, resulting in an alignment score of 1, indicating a moderate relation. \gls{lime} highlighted words in orange, representing terms contributing to this moderate alignment. The prediction probability was 1.00 for score 1, accurately capturing this level of relation. No green-highlighted terms appeared, as there was no indication of a stronger alignment. The second plot of Figure \ref{fig:lime} shows a strong prediction for this score, aligning with the intended label.

\textbf{Higher Moderate Relation Case (Alignment Score 2):} In the third test, the \gls{co} was aligned with another \gls{po}, producing an alignment score of 2, indicating a higher level of moderate relation. \gls{lime} highlighted key terms in green, representing words that strengthened the alignment. The model predicted a probability of 1.00 for score 2, correctly reflecting this higher moderate relation. The third plot of Figure \ref{fig:lime} illustrates this accurate prediction.

\textbf{Strong Relation Case (Alignment Score 3):}In the final test, the \gls{co} was aligned with a different \gls{po}, yielding an alignment score of 3, signifying a strong semantic relationship. The prediction probability for score 3 was 1.00, with terms highlighted in green, showing their strong contribution to this high alignment. The fourth plot of Figure \ref{fig:lime} confirms the model's perfect alignment with the intended label.

These experiments demonstrate that \gls{lime} can effectively interpret model predictions for \gls{cam} automation, providing high interpretability and assurance in the model's ability to assess and score \gls{co}-\gls{po}/\gls{pso} alignment, thereby enhancing the reliability of the automated system and supporting its use in curriculum design.

\section{Conclusion}\label{sec5}
This paper presented a novel approach for automating the alignment of \gls{co} with \gls{po} and \gls{pso} using BERT-based models in conjunction with transfer learning techniques. We employed four pretrained BERT models: BERT Base, DistilBERT, ALBERT, and RoBERTa to predict the alignment score between \gls{co} and \gls{po}/\gls{pso} pairs. The performance of these BERT models was benchmarked against traditional machine learning classifiers such as Decision Tree, Random Forest, and XGBoost. The BERT-based models outperformed the traditional machine learning classifiers, demonstrating the potential of transfer learning in accurately assessing \gls{co} and \gls{po}/\gls{pso} alignment.

Data augmentation was also crucial in enhancing the robustness of the models. We implemented an augmentation strategy that replaced 30\% of randomly selected words in each \gls{co}, \gls{po}, and \gls{pso} description with synonyms. This approach helped the models generalize better by adapting to variations in phrasing and terminology, ultimately improving their predictive ability.

To enhance the transparency of the models’ decision-making process, we applied an Explainable AI technique, specifically \gls{lime}. This method provided valuable insights into the factors influencing the alignment score predictions, allowing for better interpretability of the model’s behavior. By visualizing the contributions of various input features, these techniques made the model’s decisions more understandable, which is essential for ensuring the trust and reliability of the automated \gls{cam} construction process in educational settings.

Future directions will focus on developing an end-to-end system for automated \gls{co}-\gls{po}/\gls{pso} alignment. This system will integrate the methods and findings from this study into a complete workflow that can support various educational institutions in aligning their curricula more efficiently. By automating the creation of \gls{cam} and integrating them into broader curriculum management systems, this research aims to improve educational effectiveness and facilitate data-driven decision-making in academic institutions.

\printglossary[type=\acronymtype]
\printglossary

\bibliography{sn-bibliography}
\end{document}